# PARTIAL MATCHING FACE RECOGNITION METHOD FOR REHABILITATION NURSING ROBOTS BEDS

Dongmei LIANG[1] and Wushan CHENG[2]

[1,2]College of Mechanical Engineering, Shanghai University of Engineering Science,
Shanghai 201620, China
[1]meimeimeildm@163.com
[2]cwushan@163.com

## ABSTRACT

*In order to establish face recognition system in rehabilitation nursing robots beds and achieve real-time monitor the patient on the bed. We propose a face recognition method based on partial matching Hu moments which apply for rehabilitation nursing robots beds. Firstly we using Haar classifier to detect human faces automatically in dynamic video frames. Secondly we using Otsu threshold method to extract facial features (eyebrows, eyes, mouth) in the face image and its Hu moments. Finally, we using Hu moment feature set to achieve the automatic face recognition. Experimental results show that this method can efficiently identify face in a dynamic video and it has high practical value (the accuracy rate is 91% and the average recognition time is 4.3s).*

## 1. INTRODUCTION

Right now the aging population phenomenon is more and more serious, the proportion of disability and semi-disabled elderly risen and the utilization rate of rehabilitation nursing robots beds in family and medical institutions have also increased significantly. In addition, there are differences in nursing quality, professional level and responsibility. How to implement real-time monitoring of the patient, and avoid unexpected situations have become important issues facing families and caregivers. If you add a face recognition system in rehabilitation nursing robots beds. Not only can providing patients with real-time status to family members or caregivers , standardizing caregivers behaviors, avoiding the occurrence of adverse events such as abuse of the elderly and so on, but also can improving the efficiency of health care workers and medical institutions (Such as achieved multiple wards, centralized management of multiple beds, distributed control).

Face recognition is pattern recognition technology which extract feature and recognition in Face image by computer. It includes three aspects, face detection, face representation and face recognition. The first is the human face detection and location. It is to find the position of the face exists from the input image. The face is divided from the background (face detection). Second is feature extraction and recognition in normalized face image. This is the face representation and face recognition.

Face detection method have four categories. There are knowledge-based methods, invariant feature-based methods, based on template matching methods and Appearance-based methods. Yang et al [1] using a face detection method based on a hierarchical knowledge. It is said that is the mosaic method. Dai et al [2] proposed automatic face detection method based on skin color texture features. Yuille et al [3] proposed a deformable template method used to implement face detection. Appearance-based method also includes feature face method (Eigenfaces) [4] and neural network method [5].

Currently, many countries carried out research related to face recognition. In the field of face recognition, they have formed following research directions in international. The first method is based on geometric features [6, 7]. The main representative is MIT's Brunelli and Poggio team. They used improved integral projection method to extract characterized by Euclidean distance of 35 3D face feature vector used for pattern classification. The second method is based on template matching. The main representative is Yuille of Harvard Smith-Kettlewell Eye Research Center [8]. He extracted the contour of the eyes and mouth by elastic template. The third method is based on KL transform Eigenface method [9, 10]. The main principal investigator is Pentland of MIT Media Lab. The forth method is based on Hidden Markov Model [12]. The main representative is Samaria group of Cambridge University and Nefian group of Georgia Institute of Technology. The fifth method is based on neural network. The main representative is Poggio team put forward for HyperBF neural network recognition method and the Buxton and Howell group of Sussex University put forward for RBF network identification method [11].Face recognition method which based on geometric features is the earliest and commonly used. This method utilizes the structural characteristics of facial geometry. First, we locate the main organs of the face. The main organs contain eyes, eyebrows, nose, mouth, chin and the outer contour of the face. Then we select a set of characteristic structural properties which can characterize distance, angle and area shape information between the organs. Finally, we use the distance and the ratio of other parameters as the identification feature information to identify the front and side faces. As face recognition system of the rehabilitation nursing robots beds, it must have both efficiency and accuracy. We propose a face recognition method based on partial matching which apply for rehabilitation nursing robots beds. The method using Haar classifier to detect human faces automatically in dynamic video frames, and extract facial features Hu moments. Then we using Hu moment feature set to achieve the automatic face recognition and real-time tracking.

## 2. METHOD

In the method ,firstly we using Haar classifier to detect human faces Automatically in dynamic video frames; Secondly we using Otsu threshold method to extract facial features( eyebrows, eyes, mouth) in the face image and its Hu moments; Finally, we using Hu moment feature set which contains 35 Hu Moment to achieve the automatic face recognition.

## 2.1. Face Detection

Face detection is the basis for face recognition. As a result, the accuracy of face recognition plays an important role. Face detection in rehabilitation nursing robots beds is using Adaboost algorithm to train Haar classifier [13] for face detection in dynamic video frame. Then Judge the video whether contains a face. Finally give a human face location and size information. Specific detection process shown in FIG. 1. (FIG. 1 (b) in red box marked for the detection of the face).

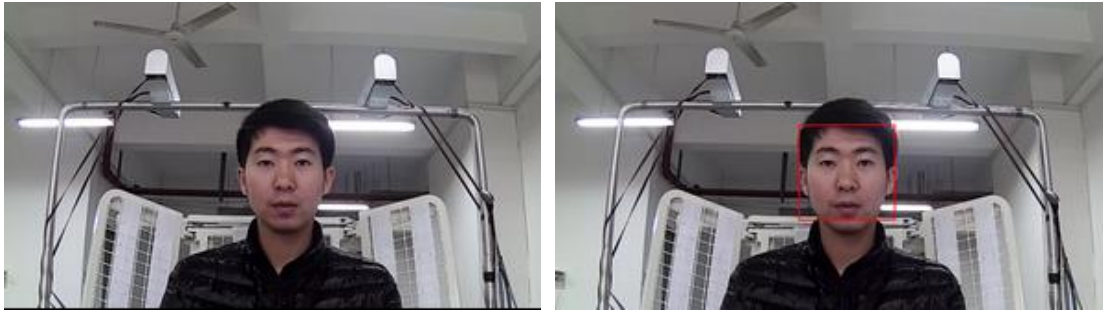

(a)  Dynamic video in a frame image  (b)  Face detection results

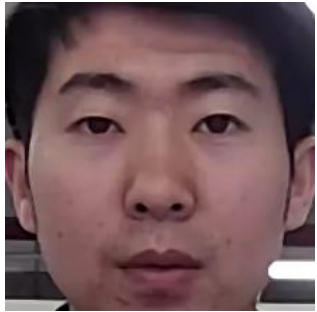

(c)  Face image
Figure 1:  Face detection process and results

## 2.2. Local Match Face Recognition Based on Hu Moments

Hu moments does not change with the position and orientation of the image. This properties is a very effective tool for extract morphological characteristics of image. On the basis of Hu, R.Y.Wong gives the moments of the calculation method of discrete state [14]. The two-dimensional function $f(x, y) \in L(R2)$ for $O - xy$ plane, the definition of $(p + q)$ order origin moment $m_{pq}$ and $\mu_{pq}$ in discrete state show in the formula (1), the formula (2).

$$m_{oq} = \sum_{x=1}^{M} \sum_{y=1}^{N} x^p y^q f(x, y) \quad (p, q = 0,1,2 \cdots) \tag{1}$$

$$\mu_{oq} = \sum_{x=1}^{M} \sum_{y=1}^{N} (x - \bar{x})^p (y - \bar{y})^q f(x, y) \quad (p, q = 0,1,2 \cdots) \qquad (2)$$

In formula, $M$、$N$ is the number of rows and columns of the image, $x$, $y$ is the coordinates of pixel of image, $\bar{x}$、$\bar{y}$ is the coordinates of center point of image.

Hu proposed seven invariant moments. It satisfies the conditions of translation and rotation invariant, show in the formula (3).

$$\begin{aligned}
Hu[0] &= \eta_{20} + \eta_{02} \\
Hu[1] &= (\eta_{20} - \eta_{02})^2 + 4\eta_{11} \\
Hu[2] &= (\eta_{30} + 3\eta_{12})^2 + (\eta_{21} + 3\eta_{03})^2 \\
Hu[3] &= (\eta_{30} + \eta_{12})^2 + (\eta_{21} + \eta_{03})^2 \\
Hu[4] &= (\eta_{30} - 3\eta_{12})(\eta_{30} + \eta_{03})[(\eta_{30} + \eta_{12})^2 - \\
&\quad 3(\eta_{21} + \eta_{03})^2] + (3\eta_{21} - \eta_{03})(\eta_{21} + \eta_{03}) \\
&\quad [3(\eta_{30} + \eta_{12})^2 - (\eta_{21} + \eta_{03})^2] \\
Hu[5] &= (\eta_{20} + \eta_{02})[(\eta_{30} + \eta_{12})^2 - (\eta_{21} + \eta_{03})^2 + \\
&\quad 4\eta_{11}(\eta_{30} + \eta_{12})(\eta_{21} + \eta_{03})] \\
Hu[6] &= (3\eta_{21} - \eta_{30})(\eta_{30} + \eta_{12})[(\eta_{30} + \eta_{12})^2 - \\
&\quad 3(\eta_{21} + \eta_{03})^2] + (3\eta_{12} - \eta_{30})(\eta_{21} + \eta_{03}) \\
&\quad [3(\eta_{30} + \eta_{12})^2 - (\eta_{21} + \eta_{03})^2]
\end{aligned} \qquad (3)$$

$\eta_{pq}$ is $f(x, y)$ normalized of $(p + q)$ order central moments, from $Hu[0]$ to $Hu[6]$ is seven Hu moments characteristic values of the variables.

We use Otsu threshold method to get binary face image and extract the facial features area. (The segmentation results of Figure 1 (c) shown in Figure 2). Then calculate seven Hu moments eigenvalues of these five parts. Hu moments eigenvalues of these five parts of Figure 2 show in Table 1.

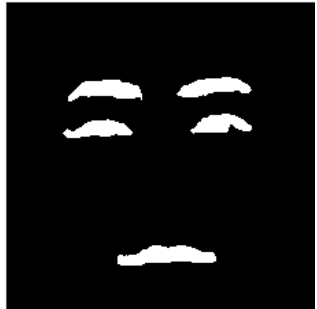

Figure 2: Face image segmentation results

Table 1: Hu moments extract eigenvalues in facial features binary image

| Hu moments eigenvalues | Left eyebrow | Right eyebrow | Left eye | Right eye | Lip |
|---|---|---|---|---|---|
| $H[0]$ | 0.309422 | 0.27892 | 0.46138 | 0.396682 | 0.414793 |
| $H[1]$ | 0.068286 | 0.049427 | 0.181526 | 0.124126 | 0.138669 |
| $H[2]$ | 0.001145 | 0.00178 | 0.005953 | 0.005539 | 0.005166 |
| $H[3]$ | 0.000133 | 0.000161 | 0.001035 | 0.00072 | 0.00088 |
| $H[4]$ | -3.00773 | -6.61529 | -5.46412 | -4.47711 | -7.74619 |
| $H[5]$ | -2.0013 | -3.10648 | -7.25124 | -0.00014 | -7.22877 |
| $H[6]$ | 4.905664 | -5.41296 | 2.363238 | 1.379213 | -1.82417 |

## 3. EXPERIMENTAL RESULTS AND ANALYSIS

### 3.1. Materials and Equipment

The face recognition system in rehabilitation nursing robots beds use Hikvision IR IP Camera (DS-2CD3Q10FD-IW)to real-time monitor the patient on the bed. The experimental System Configuration: Intel(R) Core(TM) i3 CPU, 2.00GB RAM.

The software environment: Visual c ++ 6.0.

### 3.2. Experimental Results and Analysis and Equipment

We use Hikvision IR IP Camera to test 100 times for 10 Experimenter. Experiments failed nine times in total. There are two times by the face detection error. Specific recognition results shown in Table 2（The time required to complete a face detection and recognition process is called recognition time）.

Table 2 Face recognition results

| The total number of experiments | Error Number | Accuracy rate /% | Recognition time /s |
|---|---|---|---|
| 100 | 9 | 91% | 4.3 |

According to Table 2, we using Hu moment feature set to achieve the automatic face recognition. Experimental results show that this method can efficiently identify face in a dynamic video and it has high practical value.

Rehabilitation care robots bed does not use many mature face recognition algorithm, but rather proposes a new simple and fast face recognition method. It mainly for consider the balance between efficiency and recognition rate. The process of automatic face detection has cost a certain time(The average time of detect a face require 1.1s）. In the face recognition part, sophisticated recognition algorithm can improve the accuracy .But it will sacrifices recognition efficiency. Even if the recognition rate increase to 100%. It has little significance for practical applications. Therefore, the partial matching face recognition method for rehabilitation nursing robots beds has high practical value.

## 4. Conclusion

Join intelligent system in rehabilitation nursing robots beds. Especially face recognition system can effectively solve the aging population brings serious issues. Based on this, we propose a face recognition method based on partial matching Hu moments which apply for rehabilitation nursing robots beds. This method makes full use of the geometric characteristics of the human face facial characterization capabilities. On the basis of using Haar classifier to achieve automatic detection human face on dynamic video frame. We use Hu moments feature set for automatic identification and real-time tracking for acquired face image. The method can effectively identify the face in dynamic video. When a face has expression, the face recognition rate will decline. Therefore we continue to conduct research on facial expression recognition. We combine the results of face recognition with the Control Systems of rehabilitation nursing robots beds, and further to improve the rehabilitation nursing robots beds intelligence and practicality.

## Acknowledgements

This work is supported by the Shanghai Science and Technology Committee.

**Authors**

Dongmei LIANG was born in 1990, and now the Shanghai University of Engineering Science postgraduate. Her present research interest is image analysis and processing.

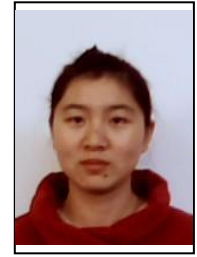